\documentclass[10pt]{article}

\twocolumn

\usepackage[utf8]{inputenc}

\usepackage{natbib}
\usepackage[acronym,nomain,nonumberlist]{glossaries} 
\makeglossaries

\usepackage[title]{appendix}
\usepackage{graphicx}
\usepackage{hyperref}
\usepackage{microtype}
\usepackage[british]{babel}
\usepackage{comment}
\usepackage{longtable}
\usepackage{booktabs}
\usepackage[shortlabels]{enumitem}
\usepackage{fullpage,amsfonts,bm}
\usepackage{amsmath,amssymb,amsthm,mathtools,bbm,physics,algorithm}
\usepackage{comment}
\usepackage[noend]{algpseudocode}
\usepackage{mleftright}\mleftright
\usepackage{hyperref} 
\usepackage{algpseudocode}
\usepackage{float}
\usepackage{authblk}
\usepackage{comment}
\floatstyle{boxed}
\makeatletter
\def\BState{\State\hskip-\ALG@thistlm}
\makeatother
\newfloat{algorithm}{t}{lop}
\floatname{algorithm}{Algorithm}

\def\01{\{0,1\}}

\title{A pragmatic approach to estimating average treatment effects from EHR data:  the effect of prone positioning on mechanically ventilated COVID-19 patients}

\author[1]{Adam Izdebski}
\affil[1]{Pacmed, Amsterdam, The Netherlands} 

\author[2]{Patrick J. Thoral, MD, EDIC}
\affil[2]{Department of Intensive Care Medicine, Laboratory for Critical Care Computational Intelligence, Amsterdam Medical Data Science, Amsterdam UMC, Vrije Universiteit, Amsterdam, The Netherlands} 

\author[1]{Robbert C.A. Lalisang, MD}

\author[3]{Dean M. McHugh}
\affil[3]{Institute of Logic, Language, and Computation, University of Amsterdam, The Netherlands} 

\author[4]{Diederik Gommers, MD, PhD}
\affil[4]{Department of Intensive Care, Erasmus Medical Center, Rotterdam, The Netherlands} 

\author[5]{Olaf L. Cremer, MD, PhD}
\affil[5]{Intensive Care, UMC Utrecht, Utrecht, The Netherlands} 

\author[6]{Rob J. Bosman, MD}
\affil[6]{ICU, OLVG, Amsterdam, The Netherlands} 

\author[7]{Sander Rigter, MD}
\affil[7]{Department of Anesthesiology and Intensive Care, St. Antonius Hospital, Nieuwegein, The Netherlands} 

\author[8]{Evert-Jan Wils, MD, PhD}
\affil[8]{Department of Intensive Care, Franciscus Gasthuis \& Vlietland, Rotterdam, The Netherlands} 

\author[9]{Tim Frenzel, MD, PhD}
\affil[9]{Department of Intensive Care Medicine, Radboud University Medical Center, Nijmegen, The Netherlands} 

\author[10]{Dave A. Dongelmans, MD, PhD}
\affil[10]{Department of Intensive Care Medicine, Amsterdam UMC, Amsterdam, The Netherlands} 

\author[11]{Remko de Jong, MD}
\affil[11]{Intensive Care, Bovenij Ziekenhuis, Amsterdam, The Netherlands} 

\author[12]{Marco A.A. Peters, MD}
\affil[12]{Intensive Care, Canisius Wilhelmina Ziekenhuis, Nijmegen, The Netherlands} 

\author[13]{Marlijn J.A Kamps, MD}
\affil[13]{Intensive Care, Catharina Ziekenhuis Eindhoven, Eindhoven, The Netherlands} 

\author[14]{Dharmanand Ramnarain, MD}
\affil[14]{Department of Intensive Care, ETZ Tilburg, Tilburg, The Netherlands} 

\author[15]{Ralph Nowitzky, MD}
\affil[15]{Intensive Care, HagaZiekenhuis, Den Haag, The Netherlands}

\author[16]{Fleur G.C.A. Nooteboom, MD}
\affil[16]{Intensive Care, Laurentius Ziekenhuis, Roermond, The Netherlands} 

\author[17]{Wouter de Ruijter, MD, PhD}
\affil[17]{Department of Intensive Care Medicine, Northwest Clinics, Alkmaar, The Netherlands} 

\author[18]{Louise C. Urlings-Strop, MD, PhD}
\affil[18]{Intensive Care, Reinier de Graaf Gasthuis, Delft, The Netherlands}

\author[19]{Ellen G.M. Smit, MD}
\affil[19]{Intensive Care, Spaarne Gasthuis, Haarlem en Hoofddorp, The Netherlands} 

\author[20]{D. Jannet Mehagnoul-Schipper, MD, PhD}
\affil[20]{Intensive Care, VieCuri Medisch Centrum, Venlo, The Netherlands} 

\author[21]{Tom Dormans, MD, PhD}
\affil[21]{Intensive care, Zuyderland MC, Heerlen, The Netherlands} 

\author[22]{Cornelis P.C. de Jager, MD, PhD}
\affil[22]{Department of Intensive Care, Jeroen Bosch Ziekenhuis, Den Bosch, The Netherlands} 
\author[23]{Stefaan H.A. Hendriks, MD}
\affil[23]{Intensive Care, Albert Schweitzerziekenhuis, Dordrecht, The Netherlands} 
\author[24]{Sefanja Achterberg, MD, PhD}
\affil[24]{ICU, Haaglanden Medisch Centrum, Den Haag, The Netherlands}

\author[25]{Evelien Oostdijk, MD, PhD}
\affil[25]{ICU, Maasstad Ziekenhuis Rotterdam, Rotterdam, The Netherlands}

\author[26]{Auke C. Reidinga, MD}
\affil[26]{ICU, SEH, BWC, Martiniziekenhuis, Groningen, The Netherlands}

\author[27]{Barbara Festen-Spanjer, MD}
\affil[27]{Intensive Care, Ziekenhuis Gelderse Vallei, Ede, The Netherlands}

\author[28]{Gert B. Brunnekreef, MD}
\affil[28]{Department of Intensive Care, Ziekenhuisgroep Twente, Almelo, The Netherlands} 

\author[29]{Alexander D. Cornet, MD, PhD}
\affil[29]{FRCP, Department of Intensive Care, Medisch Spectrum Twente, Enschede, The Netherlands} 

\author[30]{Walter van den Tempel, MD}
\affil[30]{Department of Intensive Care, Ikazia Ziekenhuis Rotterdam, Rotterdam, The Netherlands} 

\author[31]{Age D. Boelens, MD}
\affil[31]{Anesthesiology, Antonius Ziekenhuis Sneek, Sneek, The Netherlands} 

\author[32]{Peter Koetsier, MD}
\affil[32]{Intensive Care, Medisch Centrum Leeuwarden, Leeuwarden, The Netherlands} 

\author[33]{Judith Lens, MD}
\affil[33]{ICU, IJsselland Ziekenhuis, Capelle aan den IJssel, The Netherlands}

\author[34]{Harald J. Faber, MD}
\affil[34]{ICU, WZA, Assen, The Netherlands} 

\author[35]{A. Karakus, MD}
\affil[35]{Department of Intensive Care, Diakonessenhuis Hospital, Utrecht, The Netherlands} 

\author[36]{Robert Entjes, MD}
\affil[36]{Department of Intensive Care, Adrz, Goes, The Netherlands}

\author[37]{Paul de Jong, MD}
\affil[37]{Department of Anesthesia and Intensive Care, Slingeland Ziekenhuis, Doetinchem, The Netherlands}

\author[38]{Thijs C.D. Rettig, MD, PhD}
\affil[38]{Department of Anesthesiology, Intensive Care and Pain Medicine, Amphia Ziekenhuis, Breda, The Netherlands} 

\author[39]{Sesmu Arbous, MD, PhD}
\affil[39]{Intensivist, LUMC, Leiden, The Netherlands}

\author[2]{Lucas M. Fleuren, MD}
\author[2]{Tariq A. Dam, MD}
\author[1]{Michele Tonutti, MRes} 
\author[1]{Daan P. de Bruin}
\author[2]{Paul W.G. Elbers, MD, PhD, EDIC}
\author[1]{Giovanni Cinà, PhD}

\begin{document}

\begin{titlepage}
\maketitle
\end{titlepage}

\begin{center}
\begin{minipage}{\dimexpr\paperwidth-14cm}
\begin{abstract}
Despite the recent progress in the field of causal inference, to date there is no agreed upon methodology to glean treatment effect estimation from observational data. The consequence on clinical practice is that, when lacking  results from a randomized trial, medical personnel is left without guidance on what seems to be effective in a real-world scenario.

This article proposes a pragmatic methodology to obtain preliminary but robust estimation of treatment effect from observational studies,  to provide front-line clinicians with a degree of confidence in their treatment strategy. Our study design is applied to an open problem, the estimation of treatment effect of the proning maneuver on COVID-19 Intensive Care patients.

\vspace{0.5cm}


\end{abstract}
\end{minipage}
\end{center}

\section{Introduction}
\label{sec:intro}

A central problem of causal inference is estimating treatment effects from observational data. A vast body of literature has addressed the issue of finding the best method for this task, but there is no established winner \citep{dorie2019automated}. Most model comparisons are performed on synthetic data, and the race for developing new methodologies is ongoing \citep{bica2021real}.

However, clinicians in practice are often faced with the problem of not having guidance on which treatment to use. A clear example of this phenomenon is the COVID-19 pandemic, during which clinical staff had to treat patients with a novel condition for which no randomized controlled trials (RCTs) on treatment effectiveness were available \citep{guerin2020prone}.
When there is no RCT data, clinicians are left to improvise based only on their clinical experience. This points to the need for a methodology to achieve preliminary yet robust conclusions from observational data, to guide clinicians' behaviour while randomized trials are ongoing and to determine future directions for RCTs. This need is highlighted by the increasing attention that regulatory bodies in the US and Europe are placing on the use of observational evidence \citep{cave2019real, fda2019}.

In this paper we propose a procedure to arrive at sensible treatment effect estimation from Electronic Health Records (EHRs). Our proposal is twofold. 
First, we suggest to emulate a target RCT, whose results provide a meaningful point of reference \citep{hernan2016using, GERSHMAN2018946}. This is achieved by carefully replicating as much as possible the design of the past trial, i.e. using the same inclusion criteria, the same baseline covariates, and so forth. From a heuristic standpoint, this past RCT can act as a `prior', indicating how much the results obtained from EHR data diverge from what was previously known.
Second, we put forward a shortlist of methods to estimate treatment effect, including well-established ones and more recent Machine Learning techniques, with different theoretical guarantees. More than finding a single best-performing model, we intend to measure the level of agreement between the candidate models.

If our procedure is successful in both respects, meaning one can successfully emulate the design of similar RCTs from the past and the array of models provide relatively uniform estimations, we argue that the conclusions can be regarded as robust enough to guide clinical practice. This approach is `pragmatic' in the sense that it can provide sought-after preliminary knowledge when an RCT is not available.
To showcase the usefulness and applicability of this strategy, we test it on a real-world, large-scale use case, the estimation of the effect of proning on severely hypoxemic mechanically ventilated COVID-19 patients in the Intensive Care unit (ICU). The code to implement this strategy and to obtain the results described in the paper is fully available online.

\subsection{Contributions}
Our contributions can be summarized as follows:
\begin{enumerate}
    \item We implement an open-source data processing pipeline
    that integrates unstructured and noisy data from 25 Dutch intensive care units that shared data within the Dutch Data Warehouse (DDW) collaboration \citep{fleuren2021ddw}.
    \item We use the data to design an observation study that emulates the PROSEVA trial. This facilitates validity of the used causal inference methods \citep{hernan2016using}, allows for accurate treatment effect estimation \citep{forbes2020benchmarking, matthews2021comparing} and puts our estimates in a context.
    \item We found that applying prone positioning  improved the P/F ratio, with estimates varying between  14.54 and 20.11 mm Hg (depending on the model) in the time window 2-8 hours after proning and between 13.53 and 15.26 mm Hg in the time window 12-24 hours after proning (see Table \ref{tbl:outcome}). This aligns with results reported by previous RCTs and suggests a positive effect of prone positioning on patients with COVID-19 induced ARDS.
\end{enumerate}

\section{Related work}

Prone positioning, consisting of turning the patient from the supine to prone position, is a commonly used technique for the treatment of severely hypoxemic mechanically ventilated patients with acute respiratory distress syndrome (ARDS). 
The current definition of ARDS uses the P/F
ratio to classify disease severity. 
This ratio, defined as the partial pressure of O$_2$ in an arterial blood sample ($\text{PaO}_2$) divided by the fraction of inspired oxygen ($\text{FiO}_2$) ($\frac{\text{PaO}_2}{\text{FiO}_2} \text{ in mm Hg}$), subdivides patients in mild, moderate, and severe subgroups, using upper limits of 300, 200, and 100 mm Hg
respectively \citep{force2012acute}. 

\subsection{RCTs on prone positioning}
Meta-analysis of eight RCTs reported that prone positioning applied for at least 12 hours a day improved the P/F ratio on average by 23 mm Hg (95\% CI: 12, 35) as compared to the supine (not-proned, control) group \citep{munshi2017prone}.
In the PROSEVA trial, it was found that applying prone positioning for an average of 17 hours a day improved the P/F ratio on average by 15 mm Hg (95\% CI: 3, 27) \citep{guerin2013prone}
and therefore it is currently considered a cornerstone in the treatment of patients with ARDS and a P/F ratio of \textless 150 mm Hg \citep{guerin2020prone}.

Prone positioning has been used widely as adjuvant therapy for respiratory failure in severe COVID-19 induced pneumonia similar to ARDS. 
However, increasing evidence suggests that mechanisms underlying COVID-19 respiratory failure may be different from `classical' ARDS due to the frequent occurrence of pulmonary thrombosis. 
At the time of writing, no RCT has been conducted on the effect of proning on intubated patients with COVID-19 induced ARDS; furthermore, the available observational studies were conducted only on relatively small and single-center cohorts. 



\subsection{Emulating a target trial}

Given the existence of several RCTs on prone positioning, we design our observational study in order to facilitate a comparison with those randomized experiments, as well as to ensure the validity of causal inference assumptions. We note that by emulating a target trial high quality electronic health record data may be used to attempt to answer the clinical question of interest instead. Emulating a target trial requires specifying the protocol of a target trial including e.g. the inclusion criteria, treatment strategies, assignment procedures, outcome of interest, treatment effect of interest, together with a synchronization of treatment assignment and eligibility determination at time zero. Such emulation is consistent with a formal counterfactual theory of causality and allows for accurate treatment effect estimation \citep{hernan2016using, garcia2017value, GERSHMAN2018946, forbes2020benchmarking,  matthews2021comparing}.

\subsection{Estimating treatment effects from electronic health record data}

Estimating treatment effects from observational data is a fundamental problem transcending disciplinary boundaries.\ While multiple methods and strategies to arrive at causal quantities of interests do exist \citep{bica2021real}, the task itself is claimed to be notoriously hard. A typical testing ground for treatment effect estimators is synthetic data, i.e. a simulation experiment where the ground truth of the causal effect is known, which is not the case in real-world scenarios making it hard to benchmark and compare different methods on observational studies. The quest for validating causal inference methods, when ground truth causal effects are unknown, is ongoing \citep{neal2020realcause, schuler2018comparison}. Yet, there seems to be no absolute winner and the best methodology appears to be context-dependent. In this article we take a selection of
standard as well as more recent methods and put them to the test on a topical problem,
namely the estimation of the effect of proning on COVID-19 patients. 
We use them to arrive at clinically meaningful conclusions from unprocessed, unstructured and highly noisy EHR data. To our best knowledge our study is the first large-scale observational study investigating the effect of prone positioning on patients with COVID-19 induced ARDS providing valuable guidance for clinicians.

\section{Study design}
\label{section:cohort}

We carefully designed our data set to emulate an RCT, in order to provide the most meaningful term of comparison and ensure validity of our study. Emulating an RCT means using the same inclusion criteria, selecting the same covariates, measuring a similar outcome, and so forth. This strategy allowed us to compare ATE estimates with the literature, and we were able to achieve a reasonable level of confidence in the fulfillment of the identifiability assumptions, which are not testable from observational data (see Appendix \ref{section:causalinference} for a discussion).

We note that, since COVID-19 is a new disease, emulating an RCT does not mean replicating it. As far as we know, the treatment effect of proning for COVID-19 ARDS could be different from the one for standard ARDS, so the PROSEVA RCT should not be regarded as providing the ``true'' effects that the models are trying to estimate. Nonetheless, the two pathologies are still close enough that the PROSEVA trial can provide a meaningful term of comparison; indeed this similarity is why proning has been employed for  COVID-19 ARDS. For this reason the PROSEVA trial can i) provide a blueprint for the kind of experiment we want to run for proning on COVID-19 patients -- in terms of covariates, inclusion criteria, outcome definition, etc -- and ii) serve to put the observational ATE estimates in context. 
 
\begin{figure}[htbp]
  \centering 
  \includegraphics[width=\linewidth]{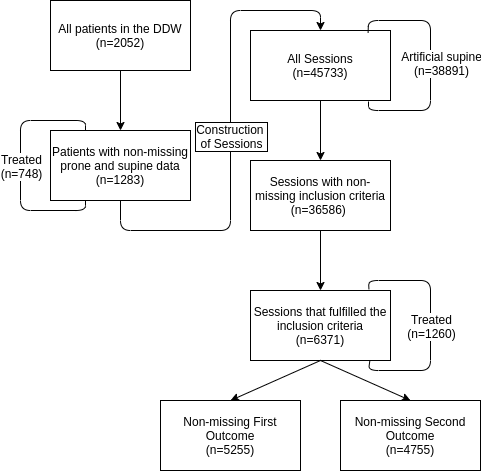}
  \caption{Overview of our study design.}
  \label{fig:flowchart} 
\end{figure} 

\subsection{Constructing Observations}

\begin{figure*}[htbp]
  \centering 
  \includegraphics[width=\textwidth]{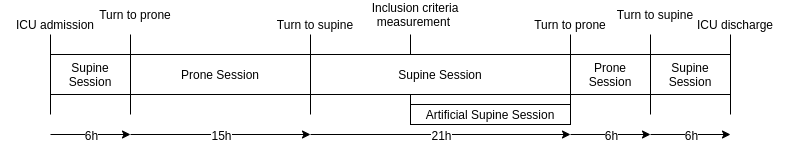}
  \caption{We describe how the data of a fictitious ICU patient would be converted into four supine and two prone sessions. An artificial supine session is created from a long supine session if there is a corresponding measurement allowing to check the inclusion criteria performed at least 8h before the end of the original session.}
  \label{fig:sessionconstruction} 
\end{figure*}

The data contained information of 2052 patients recorded in the DDW who were admitted to the ICU between March and October 2020. All patients were diagnosed with COVID-19. For all patients we extracted data about turning a patient into prone and supine position. 
The data of each patient was sliced into prone and supine sessions, where a session was defined as a time interval throughout which the patient was in the same position, either prone or supine. This is exemplified in Figure~\ref{fig:sessionconstruction}. All created sessions were later selected depending on adherence to the inclusion criteria (see Section \ref{section:inclusion}).


We took sessions to be the observations of our study, meaning that a single patient can be represented by multiple observations. A session was defined as a time interval throughout which the patient was in the same position, either prone or supine.

Sessions were employed as observations to have a data point for every moment in which a treatment decision of proning could have been made. Artificial supine sessions were created for this reason, since during a supine session there are multiple occasions in which clinicians can decide to turn the patient to prone position. 
An additional artificial supine session was created if, at least 8h before the end of the original supine session, the patient's blood gas (PaO$_2$) and ventilator values (PEEP, FiO$_2$) were recorded again.
The newly created artificial session ended at the same time point at which the original supine session ended. Altogether, we created 45733 sessions of which 2762 were prone, 4080 were original supine sessions and 38891 were artificial supine sessions.

\subsection{Constructing Features and Outcomes}

28 covariates were constructed from the data collected in the DDW. They correspond to all covariates present in the PROSEVA trial except for sepsis, McCabe and SAPS II scores, which were not available in the DDW. We included two additional covariates, one indicating morbid obesity (BMI $>$ 35) and one for driving pressure, since plateau pressure measurements were missing for the majority of observations, because most patients in Dutch ICU's were ventilated in a pressure controlled mode. 
The list of included covariates can be found in Table~\ref{tbl:tableone}. 

For each session we constructed covariates from the DDW by taking the last measurement in the eight hour interval before the start of the session. Since in practice there can be some delay in the recording of such parameters, if a measurement was missing, the first measurement was taken in the 30 minutes interval following the start of the session.

We defined two outcomes. The first outcome was the last P/F ratio measurement recorded in the time interval from two hours up to eight hours from the start of the session but before the end of the session. This outcome corresponded to the measurement taken just after the prone/supine session started (`early proning effect'). The second outcome was the last P/F ratio measurement recorded in the time interval from 12h up to 24h from the start of the session but before end of the session. Given the average length of prone sessions in this cohort ($\approx$19 hours), this outcome corresponded to the measurement taken just before the session ends (`late proning effect'). Consequently, all sessions shorter than 12h will not have a defined late proning effect.

Because of the way observations were defined, the outcomes measured the average treatment effect of a single prone session on the P/F ratio, as opposed to measuring the treatment effect of repeated proning sessions applied for a fixed time interval (e.g. three days for the PROSEVA trial). Beside the medical relevance of intermediate outcomes, this methodological choice was  due to the fact that in practice patients were not proned for an identical amount of hours for a few days in a row, so the exact administering of the proning sessions from the RCT cannot be replicated.
Furthermore, the aforementioned way of constructing covariates achieved perfect separation between covariates and outcomes, i.e. covariates were  always measured before the outcomes, allowing for valid causal inference. 

\subsection{Cohort Selection and Inclusion Criteria}\label{section:inclusion}


All sessions that did not have corresponding PaO$_2$, FiO$_2$ and PEEP baseline measurements were discarded. Since non-invasive mechanical ventilation for COVID-19 patients was rare in the Netherlands (about 1\%) \citep{fleuren2021ddw}, all sessions reporting the combination of PEEP and FiO$_2$ were considered to be sessions concerning mechanically ventilated patients. Furthermore, we discarded all proning sessions longer than 96 hours, since such sessions were extremely rare, and from a medical perspective were most likely the result of data entry errors or omissions. Following this, the inclusion criteria from the PROSEVA trial were applied for the resulting sessions. These were P/F~ratio~$<~150$~mm~Hg,
FiO$_2 \geq 60\%$ and PEEP $\geq 5$ cm of water. We chose not to include the criterion `tidal volume about 6 ml/kg', since the original PROSEVA paper did not specify explicit cutoff values for tidal volume. In addition, current Dutch guidelines for mechanical ventilation of COVID-19 patients already suggest to ventilate COVID-19 patients similarly to ARDS patients with tidal volumes of about 6 ml/kg \citep{heunksnvic2020}. For both outcomes, all observations with a missing outcome were discarded (18\% for the first and 25\% for the second outcome). Consequently, for the late proning effect we automatically discarded all sessions shorter than 12 hours, since they cannot have an outcome in the window from 12 to 24 hours. For all numerical variables we imputed data with mean values. All binary variables were imputed with the value \emph{False}. Imputation was performed in such a way to prevent any form of data leakages between data splits. 
This selection resulted in 6371 observations (from 745 patients) included in our study, 1260 (from 493 patients) of which were prone, significantly more than the PROSEVA trial that included 466 observations (each corresponding to a patient), 237 of which were prone.


\subsection{Feature Characteristics and comparison to PROSEVA Trial}
All the previous steps of our study design were defined in order to create a sub-population from the available EHR data that emulates the PROSEVA trial. The characteristics of all observations that fulfilled the inclusion criteria are summarized in Table~\ref{tbl:tableone}. 

Compared to the PROSEVA trial, the average COVID-19 patient was slightly older, with a higher incidence of diabetes, chronic renal failure and COPD. The lower average SOFA score is in line with the observation that COVID-19 patients suffer primarily from respiratory failure but not from septic shock or multiple organ dysfunction syndrome that typically accompany ARDS patients. On average, observations in our study had higher PEEP at similar or lower FiO$_2$. Tidal volume was slightly above the recommended 6 ml/kg PBW with lower respiratory rates compared to the PROSEVA ARDS patients. 

In our study higher levels of PEEP were applied and lower P/F ratios measured during prone sessions compared to PROSEVA, which may be explained by delayed proning in overwhelmed ICUs or uncertainty of effects of proning for COVID-19 patients.
The most imbalanced variables between the prone and supine sessions of our study were FiO$_2$, PaO$_2$, P/F ratio, PEEP, SOFA Score and the usage of medications. 

The PROSEVA trial similarly reported imbalance with respect to the SOFA Score and medication usage, but not w.r.t.\ FiO$_2$, P/F ratio, PEEP and PaO$_2$. 

\subsection{Data Sharing}
Within the boundaries of privacy laws and ethics, access to the DDW can be requested on the portal of the consortium: \url{www.icudata.nl}. The code used to process the data, implement the causal inference models and perform the experiments is available online.

\begin{table}[htbp]
\caption{Characteristics of observations that fulfilled the inclusion criteria. We report mean values with standard deviation for numerical variables and frequency for binary variables, ignoring missing values.
}
\label{tbl:tableone}
\begin{center}
\begin{scriptsize}
\begin{sc}
\begin{tabular}{lccc}
\toprule
Characteristic & Prone Sessions & Supine Sessions 
\\
\midrule
Age -- yr.          & $63.6 \pm 10.6$ & $63.5 \pm 10.3$  \\
Male sex -- \% & $73$\% & $77$\%  \\
BMI  & $28.3 \pm 5.2$ & $28.8 \pm 5.6$  \\ 
SOFA Score & $7.5 \pm 3.2$ & $8.0 \pm 3.4$  \\
\midrule
Diabetes -- \%                  & $25\%$ & $25\%$  \\
Renal failure -- \%             & $19\%$ & $20\%$  \\
Hepatic disease -- \%           & $5\%$  & $6\%$   \\
Coronary artery disease -- \%   & $5\%$  & $7\%$   \\
Cancer -- \%                    & $ 8\%$ & $9\%$   \\
COPD -- \%                      & $19\%$ & $16\%$  \\
Immunodeficiency -- \%          & $14\%$ & $15\%$ \\
Morbid Obesity -- \%            & $9\%$  & $13\%$ \\  
\midrule
Vasopressors -- \%                  & $64\%$ & $56\%$  \\
Neuromuscular blockers -- \%        & $46\%$ & $25\%$  \\
Renal--replacement therapy -- \%    & $7\%$  & $11\%$  \\
Glucocorticoids -- \%               & $5\%$  & $10\%$ \\
\midrule
Tidal volume -- ml                  & $454.7 \pm 124.0$ & $473.3 \pm 134.4$  \\
Tidal volume -- ml \ kg of PBW    & $6.7   \pm 1.8$   & $6.9 \pm 1.9$      \\
Respiratory rate                    & $23.2  \pm 8.7$   & $24.3 \pm 9.2$     \\
PEEP (cm H$_2$O)                    & $13.1  \pm 3.6$   & $12.4 \pm 3.7$     \\
FiO$_2$ -- \%                       & $79.0  \pm 14.8$  & $71.0 \pm 12.5$    \\
Plateau pressure -- cm H$_2$O       & $25.8  \pm 6.6$   & $26.7 \pm 6.7$     \\
Driving pressure -- cm H$_2$O       & $13.8  \pm 5.6$   & $14.0 \pm 5.5$     \\
PaO$_2$ -- mm Hg                    & $69.4  \pm 12.8$  & $71.5 \pm 11.7$    \\
P/F ratio -- mm Hg                  & $91.0  \pm 22.9$  & $103.3 \pm 22.0$   \\
PaCO$_2$ -- mm Hg                   & $56.7  \pm 15.2$  & $56.6 \pm 16.4$    \\
Arterial pH                         & $7.3   \pm 0.1$   & $7.4 \pm 0.1$      \\
Lung compliance static              & $42.7  \pm 31.0$  & $43.4 \pm 32.0$    \\
\bottomrule
\end{tabular}
\end{sc}
\end{scriptsize}
\end{center}
\vspace{-0.69cm}
\end{table}

\newpage
\section{Methods}\label{section:causalinference}


We frame our problem of estimating the average treatment effect of proning using the potential outcome framework \citep{rubin2005causal}. Let $X$ be a vector of observed \emph{covariates} (\emph{features}) and $T$ a binary \emph{treatment} variable, where $T = 1$ means treated. In the potential outcomes framework we introduce two new random variables $Y^1, Y^0$ called \emph{potential outcomes}. We call the observed potential outcome the \emph{factual outcome} $Y$ and the unobserved one the \emph{counterfactual outcome} $Y^*$. We assume that the data is generated from a fixed and unknown distribution $p(X, T, (Y^1, Y^0))$ such that the standard \emph{identifiability} criteria hold, namely consistency 
,  positivity
and ignorability 
. In our observational setting, for each sample we observe only the factual outcome, hence the data is of the form $\{(X_i, T_i, Y_i)\}_{i=1}^n$. 

We define the average treatment effect as $  \textit{ATE} = \mathbb{E}[Y^1 - Y^0]$ and the conditional average treatment effect (CATE) as $ \tau(x) = \mathbb{E}[Y^1 - Y^0 \mid X=x]$. Under the identifiability criteria we can treat observational data as a RCT on covariate values and estimate CATE from observational data as
\begin{equation*}\label{eq:cate}
    \tau(x) = \mathbb{E}[Y \mid X=x, T=1 ] - \mathbb{E}[Y \mid X=x, T=0].
\end{equation*}

\subsection{Model evaluation}\label{section:loss}
In the task of estimating the average treatment effect of proning we are interested in assessing how close our estimate $\widehat{ATE}$ is to the true effect ATE, as measured by the error $\epsilon_{ATE} = |ATE - \widehat{ATE}|$. Similarly for the conditional effect $\epsilon_{CATE}(\hat{\tau})= \mathbb{E}_x[L\big(\tau(x), \hat{\tau}(x)\big)]$,
where $L$ is an arbitrary loss function, $\hat{\tau}$ is the estimated conditional effect and $\tau$ is the true conditional effect. 
In our setting we never observe the true effect, hence we cannot use standard causal inference evaluation metrics $\epsilon_{ATE}$ and $\epsilon_{CATE}$ used on synthetic and semi-synthetic data sets like IHDP \citep{hill}. There is no widely established way to evaluate causal inference models on real-world observational data. We can however leverage two other markers to understand whether models are providing sensible estimates.

First, we compare ATEs estimated by our models to the ATE estimated by a randomized control trial that our study emulates. By emulating the RCT's design and by ensuring that the identifiability criteria are satisfied in the observational study, the estimated ATEs should be comparable to the RCT's estimate. 
Second, as the goal of causal inference methods is not only to provide an estimate close to the RCT's estimate but also to accurately predict the outcomes, we additionally evaluate the predictive performance by calculating the root-mean-squared-error (RMSE) of each method. For each model $f$, we calculate RMSE for factual outcomes $y_i$ hence $RMSE = \sqrt{\frac{1}{n}\sum_{i = 1}^n (f(x_i, t_i) - y_i)^2}$.

\subsection{Models}
In order to perform treatment effect estimation, we selected a suite of standard and well-known causal inference models that rely on a diverse set of modeling assumptions. We employed the following models to estimate the treatment effect of proning on oxygenation: 
\begin{itemize}
    \item Linear Regression (LR)
    \item Propensity score model: Doubly Robust Inverse Probability  Weighting (DR-IPW) 
    \item Propensity score model: Blocking
    \item Bayesian Additive Regression Trees (BART)
    \item Treatment-Agnostic Representation Network (TARNet)
    \item Counterfactual Regression (CfR)
\end{itemize}

For details of the models as well as justification for their choice we refer the reader to the Appendix~\ref{models}. Clarifications on the notation can be found in Appendix~\ref{notation}, while the specification on the models' hyper-parameters are reported in  Appendix \ref{algorithmicdetails}.

\section{Results}\label{section:results}

The goal of the study was to estimate the average treatment effect (ATE) of prone positioning on P/F ratio as measured after a session starts (early proning effect)
and just before an average prone session ends (late proning effect). 
For each outcome the following procedure was instated. To estimate the ATE we performed a 80/20 train/test split. Each model was trained on 100 bootstrap re-samples of the train set (we repeatedly sampled 95\% of the train set with replacements) and with each model an ATE estimate was calculated on the test set. We obtained the ATE estimate (Table~\ref{tbl:outcome}) by taking the average of ATE estimates for every bootstrap sample of the train set. The 95\% confidence intervals are calculated as bootstrap percentile intervals (as in Section~8 of \cite{wasserman2013all}). 
Prior to calculating ATE estimates, a hyper-parameter search was performed for some of the models. This search, as well as the strategy followed for the estimation of the propensity scores, is described in Appendix~\ref{algorithmicdetails}.
As we could not measure the accuracy of the estimated ATE, we evaluated the predictive performance for each model with RMSE on the test set for each of the 100 different bootstrap samples and reported the average. The 95\% confidence intervals were calculated in the same way as for the ATE estimates.

The aim of the experiments was to assess whether the different frameworks for causal inference examined in Appendix~\ref{models} provided uniform ATE estimates. Moreover, we checked whether the estimates were close to the ATE of 15 mm Hg reported by the PROSEVA trial. Comparing the models' ATE estimates with the result of the PROSEVA trial allowed us to gain further confidence in the robustness of the models' estimates.

\begin{table}[hbtp]
\label{tbl:outcome}
{\caption{Estimated early (top) and late (bottom) proning effects. ATE and RMSE are reported with mean and 95\% CIs.}}
\vspace{0.5cm}
\setlength{\tabcolsep}{3pt}
  {\begin{tabular}{lll}
  \toprule
  \bfseries Model & \bfseries ATE & \bfseries RMSE \\
  \midrule
 LR & $15.31 (11.69, 19.80)$ & $58.48 (57.91, 59.11)$ \\
 
    DR-IPW & $14.54$ ($10.29$, $19.00$) & $59.09$  ($58.15$, $60.10$)  \\
    Blocking & $15.59$ ($11.44$, $20.29$) & $60.02$ ($58.27$, $61.71$)  \\ 
    BART & $20.11$ ($14.17$, $27.50$) & $48.62$ ($45.12$, $56.81$) \\ 
    
    TARNet & $17.70$ ($8.80$, $25.60$) & $51.79$ ($50.74$, $53.53$) \\
    
    CfR & $18.14$ ($9.28$, $27.35$) & $51.82$ ($50.83$, $53.52$) \\
  \bottomrule
  \end{tabular}}
\end{table}
\begin{table}[hbtp]
\setlength{\tabcolsep}{3pt}
  {\begin{tabular}{lll}
  \toprule
  \bfseries Model & \bfseries ATE & \bfseries RMSE \\
  \midrule
  LR & $14.47 (10.34, 19.34)$ & $53.88  (53.41, 54.58)$ \\
 
    DR-IPW & $13.53$ ($8.88$, $19.45$) & $54.14$ ($53.58$, $54.96$) \\
    
    Blocking & $14.87$ ($9.39$, $19.34$) & $53.22$ ($51.88$, $55.14$) \\ 
    BART & $13.99$ ($6.70$, $20.34$) & $50.44$ ($47.35$, $55.40$) \\ 
    
    TARNet & $15.13$ ($5.66$ $25.97$) & $50.61$ ($48.80$, $52.01$) \\
    
    CfR & $15.26$ ($5.54$, $24.94$) & $50.59$ ($48.58$, $51.54$) \\
  \bottomrule
  \end{tabular}}
\end{table}

\subsection{Average Treatment Effect Estimation} 
\begin{figure}[htbp]
  \centering
  \begin{minipage}[b]{0.49\textwidth}
  \centering
    \includegraphics[width=\textwidth]{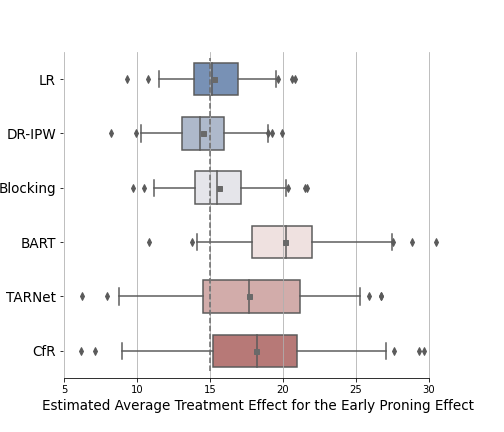}
  \end{minipage}
  \hfill
  \begin{minipage}[b]{0.49\textwidth}
    \centering
    \includegraphics[width=\textwidth]{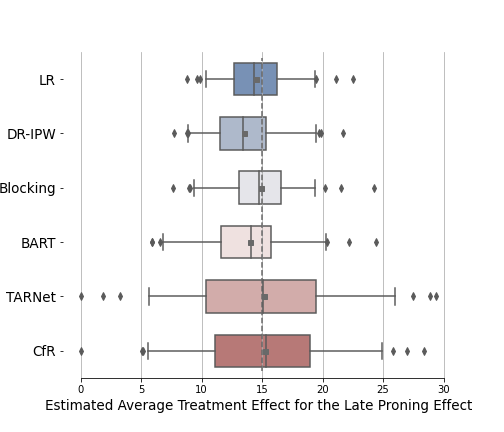}
  \end{minipage}
  \caption{Distribution of the average treatment effects estimated on the test set for the early proning effect (Up) and the late proning effect (Down). Each boxplot displays the beginning of the 95\% CI, the first quartile, the median, the third quartile and the end of the 95\% CI of all obtained ATE estimates. The black square indicates the sample mean, which is the final ATE estimate reported in Table~\ref{tbl:outcome}. The vertical, dotted line indicates the ATE from PROSEVA trial (=15).}
  \label{fig:outcome}
\end{figure}

The estimated mean ATEs ranged from $14.54$ to $20.11$ mm Hg for the early proning effect, and from $13.53$ to $15.26$ mm Hg for the late proning effect. Confidence intervals for both outcomes being strictly above zero indicated a positive treatment effect of proning on P/F ratio. 
A summary of all estimates can be found in Table~\ref{tbl:outcome} and Figure~\ref{fig:outcome}.

\subsubsection{Early Proning Effect}
The unadjusted treatment effect, meaning the  difference in the mean outcome in the treated and control group, was $12.89$ (95\% CI: $7.88$, $17.22$).\ Linear Regression (LR) and propensity score models (DR-IPW, blocking) provided the narrowest 95\% CIs for the ATE while deep learning methods (TARNet, CfR) the widest ones. This was probably due to the fact that the latter models are more prone to overfitting on the different bootstrap samples. Linear Regression, DR-IPW weighting and Blocking reported the ATE estimates closest to the PROSEVA trial, with BART, TARNet and CfR giving slightly higher estimates. 

The best performing model with respect to predictive performance was BART (RMSE: $48.62$), slightly outperforming TARNet and CfR (RMSE: $51.72$ and $51.82$ respectively). The performance of the selected models was in line with experiments on synthetic data (Table~1 in \cite{shalit2017estimating}), where BART, CfR and TARNet reported comparable errors with respect to ATE estimation ($\epsilon_{ATE}$) in out-of-sample testing on both IHDP and Jobs data sets. Linear Regression and propensity models were outperformed by more recent techniques with respect to predictive performance, probably due to the greater flexibility of the latter methods. This might explain also why the ATE estimates were slightly higher for the non-linear models.

\subsubsection{Late Proning Effect}
The unadjusted treatment effect was equal to $11.88$ ($95\%$ CI: $7.88$, $16.91$). Compared to the early proning effect, all models provided lower ATE estimates. The estimates reported for the two outcomes differed significantly, with BART reporting the wider gap ($20.11$ vs.\ $13.99$) as opposed to Linear Regression giving more similar estimates ($15.31$ vs. $14.47$). Compared to the early proning effect, all estimates were closer to the treatment effect from the PROSEVA trial. 

The predictive performance of BART, TARNet and CfR remained the strongest with similar RMSE as for the early proning effect. Obtaining ATE estimates for the late proning effect seemed to be an easier task for the Linear Regression and propensity models as they reported smaller RMSE compared to the early proning effect. While non-linear models still outperformed linear ones, the difference was much smaller in RMSE, indicating that perhaps non-linear effects played less of a role for this outcome.

%

Models of a similar kind, both Linear Regression and propensity models, and CfR and TARNet reported comparable ATE estimates and predictive performances. This was somewhat surprising, since one would expect that the correction that propensity score models and CfR provide on top of Linear Regression and TARNet (respectively) would improve their performance and possibly modify their ATE.
One of the possible explanations is that our data was rather balanced, with high treatment assignment variation. 
As there was no observable metric to measure the accuracy of CATE estimates and RCTs do not provide CATE estimates to compare with, it remains an open question to what extent these models could be used for CATE estimation. 


\section{Discussion} 





In this article we addressed an urgent problem, namely the lack of guidance for clinicians when RCT results are not available, by describing an approach to obtain information on treatment effectiveness from EHR data. 
We suggested that a multi-viewpoint approach, where different models are tested in parallel, coupled with an RCT-like experimental design can be employed to reach a reasonable level of confidence in the result of an observational study. This approach was tested on an important use case, namely treatment effect of the proning maneuver on COVID-19 patients on the ICU.
Our experiments showed that our diverse set of models reached a relatively uniform conclusion. Moreover, our findings suggested that mechanically ventilated COVID-19 patients benefit from proning to a similar extent as established by the PROSEVA trial.
This provides important medical knowledge in a situation when RCTs on proning for mechanically ventilated COVID-19 patients are not yet available and observational evidence is limited to single-center and relatively small cohorts \citep{pan2020lung, perier2020effect,berrill2021evaluation, weiss2021prone,  shelhamer2021prone}.

The extent to which results such as the ones presented here can be used to directly affect clinical decision making, or can be contrasted with RCTs, should be a matter of further discussion, as well as scrutiny from a regulatory perspective \citep{eichler2020novel}.
These results cannot be considered definitive and a randomized trial is required to formally establish the effectiveness of the treatment. Repetition of these experiments on another observational COVID-19 cohort may also help in assessing the robustness of the estimates; the fully documented and open-source code released with this article facilitates and encourages such follow-up analyses. Despite their potential brittleness and biases, observational studies such as the one presented here can help provide temporary answers for clinicians while RCTs are developed, as well as accelerate and guide prospective studies \citep{beaulieu2020examining}. Lastly, further improvement to the models' predictive performance could warrant the investigation of the effect of proning on subgroups or of the threshold on P/F ratio for which proning is beneficial.

\paragraph{Limitations}
In conclusion, our results provide further evidence on the effectiveness of proning for the treatment of mechanically ventilated COVID-19 patients. The design of our study, implemented in fully open-source code, allows for reproducibility on other COVID-19 cohorts and provides a blueprint for treatment effect estimation in scenarios where RCT data is lacking.

Regarding the limitations that should be considered when weighing the result of this study. First, we do not compare randomized experiments and observational studies with identical designs, such as in \citet{shadish2008can}, since from the DDW data it is clear that in practice the proning maneuver is not administered exactly as in the RCTs. This also holds for the outcome: the difference in treatment led us to consider oxygenation after proning, which differs from the oxygenation outcome after three days of repeated proning as defined in the PROSEVA trial.
It should be noted however that RCTs already display a great deal of variability. The proning maneuver itself is performed differently in all RCTs, and the outcomes are also measured at different time points, thus the discrepancy between our study and the RCTs is arguably not greater than the one existing between RCTs.
It also worth noting that the usage of sessions introduces dependence between samples from the same patient and might lead to over-representation of some individuals; sicker patients, for example, might stay longer in the ICU and generate more sessions. However, we do not register any additional imbalance in the distribution of baseline covariates, as reported in Table \ref{tbl:tableone}. Furthermore, we sought to mitigate this problem by employing bootstrap samples. We  note that the construction of multiple data points per patients opens the possibility to study repeated or time-varying treatment by means of causal inference methods for longitudinal data; this line of enquiry is left for future work. 

Second, all models assume the absence of unmeasured confounders. To facilitate comparison with previous RCTs, we decided to limit the set of covariates to those in the study by \citet{guerin2013prone}. While this expert-picked list ensures that most relevant variables are considered, hence giving some measure of reassurance that the assumption is fulfilled, it is already known that it might be relevant to add a few more covariates to have a more complete picture of the patient, e.g. fluid balance. 
We believe that for our case study it is possible to minimize the impact of this assumption, since the ICU routinely monitors patients in great detail. It is therefore reasonable to assume that the vast majority, if not all, of relevant parameters are recorded in EHRs, meaning that a future iteration of this study on a larger set of covariates could provide more robust results.

Third, it could be argued that a different or wider selection of models is more appropriate. The guiding principle of our choice of models was to represent both traditional methods used in the medical community as well as top performers from previous causal inference competitions, to arrive at modern Machine Learning techniques with strong theoretical guarantees. 

Finally, a further point of improvement for our approach is the enhancement of the comparison between the different models. We adopted simple and well-known metrics such as RMSE, but more advanced techniques are becoming available for observational studies, such as those by  \citet{alaa2019validating}, allowing for a more fine-grained analysis of the models' differences.


\newpage

\bibliography{bibliography.bib}

\newpage

\onecolumn

\appendix


\section{Dutch Data Warehouse}\label{DDW}

The DDW was constructed by combining, unifying, and structuring EHR data from 25 different ICUs in the Netherlands. A dedicated team of ICU clinicians, data scientists, and IT staff put great effort in the medical validation of the data by assessing data quality at several stages in the extensive data validation process \citep{fleuren2021ddw}. Since raw EHR data contains all routinely and frequently captured  clinical information, this whole endeavour eventuated in a unique dataset containing highly granular data of COVID-19 patients throughout their entire ICU stay.

The DDW includes data on demographics, comorbidities, monitoring and life support devices, laboratory results, clinical observations, medications, fluid balance, and outcomes. While the DDW is continuously growing, it contained over 400 million data points from 2052 COVID-19 patients at the moment of accessing.

\section{Notation}\label{notation}
Throughout the text we associate each observation with an index $i,j,k \in \mathbb{N}$. The covariate (feature) space $\mathcal{X} \subseteq \mathbb{R}^d$ and outcome space $\mathcal{Y} \subseteq \mathbb{R}$ are defined as usual. We denote random variables with capital letters $X, Y, Z$. We define the covariate vector $X$, outcome vector $Y$ and treatment indicator $T$ and write $X_i, Y_i, T_i$, if they correspond to an observation $i$. We denote realizations of random variables with corresponding lower case letters $x, y, z$. If a realization of a random variable $X$ is associated with the observation $i$, then we add a subscript $x_i$. When not necessary, we omit the corresponding subscripts. When a random variable $X$ follows a distribution $p$, we will write $\mathbb{E}[X]$ instead of $\mathbb{E}_{x \sim p(x)}[x]$ to denote the expected value. 

\section{Models}
\label{models}
\label{section:modes}
We choose models that rely on a diverse set of modeling assumptions, with the goal of estimating the average treatment effect of prone positioning. We introduce traditional methods for causal inference:\ linear regression and propensity score based models in Section~\ref{section:ols} and Section~\ref{section:pmodels}, the popular Bayesian regression model BART in Section~\ref{section:bart} and a recent deep-learning based framework in Section~\ref{section:cfr}. The way we choose hyper-parameters for each of the models, as well as details on implementation, is described in Appendix~\ref{algorithmicdetails}.


\subsubsection{Linear Regression}\label{section:ols}
Linear regression (LR) is the simplest parametric approach to estimating average treatment effect from observational data \citep{hernan2020causal}. 
We add it to the shortlist of models since it is widely used in the literature, and it is a valid way to adjust for confounders under the following assumption \citep{hernan2020causal}. Linear regression assumes that the conditional expectation of the outcome $Y$ can be expressed as
\begin{equation}\label{eq:ols}
    \mathbb{E}[Y \mid X=x, T=t] = \alpha_0 + \alpha_1 \cdot t + \alpha_2 \cdot x
\end{equation}
We estimate weights in Equation~\ref{eq:ols} by fitting a linear regression model and use the coefficient of the treatment effect indicator $\alpha_1$ as the estimated average treatment effect. This is because by Equation~\ref{eq:ols} it holds that 
\begin{equation*}
\begin{split}
    \tau(x) & = \mathbb{E}[Y \mid X = x, T = 1] - \mathbb{E}[Y \mid X=x, T=0] \\ & = \alpha_1
\end{split}
\end{equation*}
Therefore, not only is the outcome $Y$ a linear combination of the treatment $T$ and covariates $X$, but also the treatment effect is constant across different values of covariates $X$.
\subsubsection{Models Based on Propensity Score}\label{section:pmodels}
A propensity score, introduced by \citet{rosenbaum1983central}, is defined as the conditional probability of receiving the treatment  
$e(x) = \mathbb{P}(T = 1 \mid X = x)$.

In our setting the true propensity score is unknown and needs to be estimated from data using a \emph{propensity score model}. Such model is used to correct the linear regression model in to two ways: blocking and inverse probability weighting (IPW) \citep{imbens2004nonparametric}.


First, we use blocking (sometimes referred to as `stratification') to partition the observational data into subsets (blocks) containing observations with an estimated propensity score in a given range. We estimate the treatment effect within each block by fitting an linear regression to the data within the block. 
We obtain the final ATE estimate by taking the average of block-wise estimates, weighted by the proportion of data points in each block.

Second, following \citet{kang2007demystifying} we define a doubly robust estimator that combines inverse probability weighting with linear regression (DR-IPW) as follows. For each observation with covariate value $x$ and treatment indicator $t$, we use the estimated propensity score $\hat{e}(x)$ to define a weight \begin{equation*}\label{eq:wols}
    w(x, t) = \Big[\hat{e}(x)^t \cdot (1 - \hat{e}(x))^{(1-t)}\Big]^{-1}
\end{equation*} 
We then fit an linear regression weighted by $w$ to our data, giving more importance to observations with an unlikely treatment assignment. The ATE estimate is the treatment indicator's coefficient of the weighted model, as for the linear regression. 

The idea behind both models is to fit an linear regression to a modified version of the original data, using the propensity scores to balance the difference between the treated and control group and to correct
the fact that treatment is not independent of covariates. This can be beneficial in a variety of theoretical scenarios, for a detailed discussion of the above models and their statistical properties \citep{imbens2004nonparametric, kang2007demystifying}.

\subsubsection{BART}\label{section:bart}
Bayesian Additive Regression Trees (BART) is a nonparametric, Bayesian method for estimating functions using regression trees \citep{chipman2010bart}. It is a popular method for causal inference \citep{dorie2019automated}, used often as a baseline comparison for state-of-the-art deep learning methods. The model's popularity is based on its outstanding performance in multiple real world scenarios \citep{zhou2008extracting,blattenberger2014avalanche}, as well as simulated ones \citep{dorie2019automated}.

BART assumes that the conditional expectation of the outcome $Y$ can be expressed as 
\begin{equation*}
    \mathbb{E}[Y \mid X=x, T=t] = g(x, t) + \epsilon; \quad \epsilon \sim \mathcal{N}(0, \sigma^2)
\end{equation*}
where $g$ is an arbitrary function, $\epsilon$ is an error term and $\sigma^2$ is a constant. BART approximates function $g$ with a sum $\sum_{m=1}^M \mathcal{T}_m(x, t)$ of $M$-many regression trees $\mathcal{T}_1, \ldots, \mathcal{T}_M$. We obtain a CATE estimate by taking the difference $\hat{\tau}(x) = \sum_{m = 1}^M \big( \mathcal{T}_m(x, 1) - \mathcal{T}_m(x, 0) \big)$. The ATE estimate is obtained by averaging over CATE estimates computed on the test set $\widehat{ATE} = \frac{1}{n} \sum_{i=1}^n \hat{\tau}(x_i)$.




\subsubsection{Counterfactual Regression}\label{section:cfr}
Counterfactual Regression (CfR) is a general method for estimating CATE that is based on a theoretical framework developed by \citet{shalit2017estimating}. CfR is a regularized minimization procedure that simultaneously tries to predict the correct observed outcomes and to find a representation that balances the treated and control group.
Thus CfR fits a representation $\Phi: \mathcal{X} \xrightarrow{} \mathcal{X}$ and a regression function $f: \mathcal{X} \times \{0,1\} \xrightarrow[]{} \mathcal{Y}$, where $\mathcal{X}$ is the covariate space and $\mathcal{Y}$ is the outcome space, in order to minimize the objective
\begin{multline}\label{eq:objectivecfr}
        \min_{f, \Phi: \; ||\Phi|| = 1} \frac{1}{n} \sum_{i=1}^n w_i \cdot (f(\Phi(x_i), t_i) - y_i)^2  + \lambda \cdot \mathcal{R}(f)  + \alpha \cdot \textit{Wass}\Big(\{ \Phi(x_i) \}_{t_i = 1}, \{ \Phi(x_i) \}_{t_i = 0}\Big)
\end{multline}
where $w_i = \frac{1}{2}(\frac{t_i}{u} + \frac{1 - t_i}{1-u})$ is a weight accounting for the treatment imbalance, $u = \sum_{i=1}^n \frac{t_i}{n}$ is the fraction of observations being treated, $\alpha$ and $\lambda \cdot \mathcal{R}(f)$ are arbitrary regularization terms and $\textit{Wass}$ is the Wasserstein distance defined by \citet{villani2008optimal}. Theorem 3 of \citet{johansson2020generalization} guarantees that by minimizing Equation~\ref{eq:objectivecfr} we obtain $f, \Phi$ such that $\hat{\tau}(x) = f(\Phi(x), 1) - f(\Phi(x), 0)$ is a consistent CATE estimator. We also employ a special case of CfR, called TARNet, whose objective is given by Equation~\ref{eq:objectivecfr} with $\alpha = 0$. We follow CfR's and TARNet's implementation proposed by \citet{shalit2017estimating}. We use CfR and TARNet to obtain the ATE estimate by averaging over CATE estimates computed on the test set $\widehat{ATE} = \frac{1}{n} \sum_{i=1}^n \hat{\tau}(x)$. 

\section{Algorithmic Details}
\label{algorithmicdetails}

\subsubsection{Propensity Score Models}\label{propensityscoremodels}

\begin{figure}[!thp]
  \centering
  \begin{minipage}[b]{0.49\textwidth}
    \includegraphics[width=0.95\textwidth]{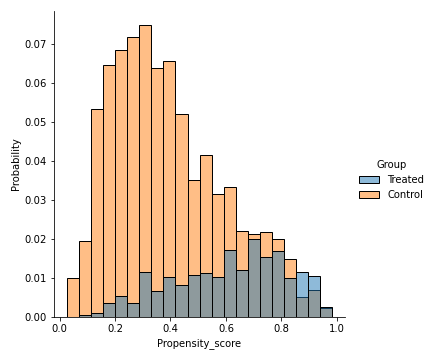}
  \end{minipage}
  \hfill
  \begin{minipage}[b]{0.49\textwidth}
    \includegraphics[width=0.95\textwidth]{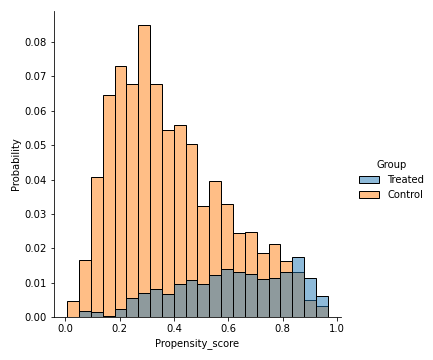}
  \end{minipage}
  \caption{Distribution of the estimated propensity scores for the data set corresponding to the early proning effect (Up) and the late proning effect (Down). The plots show a significant overlap for the estimated propensity scores above $0.1$.}
  \label{fig:pscoredits}
\end{figure}

We evaluate three different strategies to estimate the propensity score. We fit a logistic regression model to 1) confounders chosen by ICU doctors,
2) all covariates, 3) all covariates with interactions. The second model is preferable to the first as adjusting with the second model results in lower imbalance, when measured by the absolute value of the normalized mean difference defined by \citet{imbens2015causal}. We choose the second model over the third, despite a slightly lower imbalance, because the propensity scores estimated by the third model tend to extreme values, making it less useful for calculating IPW weights. We use a scikit-learn implementation for the logistic regression with no regularization and with weights adjusting for class imbalance \citep{sklearn_api}. After estimating the propensity scores (Figure~\ref{fig:pscoredits}) we perform propensity score clipping ($= 0.1$) to account for the lack of overlap in the region with values below $0.1$. For the linear regression models we use the implementation proposed in \emph{statsmodels} package \citep{seabold2010statsmodels}.

\subsubsection{BART}\label{appendix:bart}

We use BART as implemented by \citet{bartMachine}. We train BART with $m=50$ trees and use 5-fold CV (with respect to the lowest RMSE) on the train data to choose hyperparameters: $k = 2$, $\nu = 3$, $q=0.9$ controlling the model regularization. These hyperparameters are reported as default by \citet{chipman2010bart}. Additionally, we use default parameters $\alpha = 0.95, \beta = 2$ specifying the prior over trees' structure. 

\subsubsection{CFR and TARNet}\label{appendix:cfr}

We follow CfR's and TARNet's implementation proposed by \citet{shalit2017estimating}. They implement CfR and TARNet as a feed-forward neural network with three fully-connected layers and with ELU activation functions for both the representation $\Phi$ and for the regression functions $f_1 = f(x, 1)$ and $f_0 = f(x, 0)$. We use default hyperparameters with layer size of 200 for the representation and 100 for outcome networks. The model is trained using Adam optimizer and $\ell_2$ weight decay for outcome networks \citep{kingma2014adam}. We choose default regularization hyper-parameters: $\lambda = 0.0001$ and $\alpha=1$ ($\alpha=0$ for TARNet). We perform early stopping w.r.t.\ surrogate mean-squared-error defined by \citet{shalit2017estimating} and calculated on a held-out validation set. We perform a grid search evaluated on the validation set in order to access whether to use different than default hyperparameters for TARNet and CfR. We decided to use the default hyperparameters for all models as improvement w.r.t. their predictive performance was negligible.

\section{Medical Abbreviations}\label{appendix:medical}
\newacronym{covid-19}{COVID-19}{Coronavirus disease 2019}
\newacronym{icu}{ICU}{Intensive care unit}
\newacronym{ehr}{EHR}{Electronic health record}
\newacronym{rct}{RCT}{Randomized controlled trial}
\newacronym{pao2}{PaO2}{Partial pressure of O2 in an arterial blood sample}
\newacronym{fio2}{FiO2}{Fraction of inspired oxygen}
\newacronym{p/f}{P/F ratio}{Ratio between PaO2 and FiO2 (PaO2/FiO2)}
\newacronym{ards}{ARDS}{Acute respiratory distress syndrome }
\newacronym{peep}{PEEP}{Positive end-expiratory pressure}
\newacronym{copd}{COPD}{Chronic obstructive pulmonary disease}
\newacronym{sofa}{SOFA score}{Sequential organ failure assessment score}
\newacronym{bmi}{BMI}{Body mass index}
\newacronym{pbw}{PBW}{Predicted body weight}
\newacronym{bw}{BW}{Body weight}

\glsaddall
\setglossarysection{section}
\renewcommand{\glossarysection}[2][]{} 
\printglossary


\end{document}